\documentclass[conference]{IEEEtran}
\IEEEoverridecommandlockouts
\usepackage{cite}
\usepackage{amsmath,amssymb,amsfonts}
\usepackage{algorithmic}
\usepackage{graphicx}
\usepackage{textcomp}
\usepackage{xcolor}
\usepackage{subfig}
\usepackage{multirow}

\def\BibTeX{{\rm B\kern-.05em{\sc i\kern-.025em b}\kern-.08em
    T\kern-.1667em\lower.7ex\hbox{E}\kern-.125emX}}
\newcommand{\reffig}[1]{Fig. \ref{#1}}

\newcommand{\reftab}[1]{Table \ref{#1}}

\usepackage{hyperref}  %

\usepackage{fancyhdr}

\fancypagestyle{doi}{\lfoot{DOI reference number: 10.18293/SEKE2022-076}}
\begin{document}

\title{Multi-Frames Temporal Abnormal Clues\\Learning Method
	for Face Anti-Spoofing\\
}

\author{
	\IEEEauthorblockN{Heng Cong\textsuperscript{*}\thanks{* denotes corresponding author},~Rongyu Zhang, Jiarong He,~Jin Gao}
	\IEEEauthorblockA{Interactive Entertainment Group of Netease Inc, Guangzhou, China}
	\IEEEauthorblockA{\{congheng, zhangrongyu, gzhejiarong, jgao\}@corp.netease.com}
}

\maketitle
\thispagestyle{doi}

\begin{abstract}
Face anti-spooﬁng researches are widely used in face recognition and has received more attention from industry and academics. In this paper, we propose the EulerNet, a new temporal feature fusion network in which the differential filter and residual pyramid are used to extract and amplify abnormal clues from continuous frames, respectively. A lightweight sample labeling method based on face landmarks is designed to label large-scale samples at a lower cost and has better results than other methods such as 3D camera. Finally, we collect 30,000 live and spoofing samples using various mobile ends to create a dataset that replicates various forms of attacks in a real-world setting. Extensive experiments on public OULU-NPU show that our algorithm is superior to the state of art and our solution has already been deployed in real-world systems servicing millions of users.

\end{abstract}

\section{Introduction}
\label{sec:intro}

Face anti-spoofing (FAS) plays an important role in interactive AI systems and is also a challenging task without specific hardware equipped in the industry. The existing methods based on multi-modal information (e.g. infrared light, structured light, and light field) are widely used in scenarios, where there are specialized hardware devices. Although these methods \cite{liu2018learning,xie2017one,yi2014face} perform well in classification, they cannot be used on mobile devices on a broad scale.

Initially, the traditional manual feature is used to recognize face spoofing \cite{boulkenafet2015face,siddiqui2016face,li2016generalized}. With the development of convolution neural network (CNN), classification techniques are excelled in the FAS task \cite{stehouwer2020noise,lucena2017transfer,rehman2017deep}. Recently, the application of face depth information has further improved the FAS effectiveness. The previous method performed excellently, but there are many shortcomings. In \cite{yu2020auto,yu2020fas,qin2020one,qin2021meta}, single-frame information is used to estimate the face depth for video prediction, but the inter-frame information of the video is discarded. \cite{wang2020deep,asim2017cnn,yu2021transrppg,yang2019face} achieved better effects by considering the inter-frame information, whereas the feature of face abnormal clues is not be utilized. \cite{liu2019aurora} label face depth with the 3D camera equipped to focus on face detail, which is difficult to promote. By contrast, \cite{yu2020face,kim2019basn,wang2018exploiting} convert 2D faces to 3D faces using PRNet,  which has a lower label cost but a higher error rate. Especially in the prediction of large-angle side faces, \cite{yu2020face,kim2019basn,wang2018exploiting} is not effective. The binary mask label is obtained by the middle part of the face, in \cite{george2019deep,hossain2020deeppixbis,yu2020auto,ma2020novel,sun2020face}, where live and attack is marked as 1 and 0, respectively. However, \cite{george2019deep,hossain2020deeppixbis,yu2020auto,ma2020novel,sun2020face} ignores the face depth and edge texture information, which degrades the effect of the model. Moreover, many datasets such as NUAA \cite{tan2010face}, CASIA-MFSD \cite{zhang2012face}, Replay-Attack \cite{chingovska2012effectiveness}, MSU-USSA \cite{patel2016secure}, and OULU-NPU \cite{boulkenafet2017oulu} released several FAS data that are collected in the laboratory state. However, because of the uncertainty of lighting, scenarios, device, etc., the samples in the laboratory varies greatly from the samples in the real world.

To address the above problems, we propose a temporal feature fusion network named EulerNet. The eulerian video magnification is introduced to extract temporal clues and fuse the inter-frame feature from the video. Residual Pyramid is designed in EulerNet to fuse features of different resolutions. We propose a feature-compressed attention module (FCAM) to process features in the temporal dimension. To balance the cost and accuracy, we propose a lightweight face labeling method based on face landmarks, which is faster than \cite{liu2019aurora} and more accurate than \cite{yu2020face,kim2019basn,wang2018exploiting}. The face position map obtained by the proposed labeling method is easier to model at fine-grain than depth map. Compared with binary mask, more gradient information of the face edges is retained by face position map. Finally, a large and comprehensive FAS dataset is built to simulate the real usage background under the industry environment. According to the real proportion of cell phone models and application scenarios in real-world, we collect live and attack by different types of mobile ends in various scenarios for our dataset.

In summary, the main contributions of this work as follows:
\begin{itemize}
	\item We propose a novel network architecture to extract abnormal clues and process multi-frame temporal information. The feature-compressed attention module (FCAM) and Residual Pyramid are designed to improve the computational efficiency, focus on prominent face information, and amplify the weak signal.
	\item A lightweight face labeling method based on face landmarks is presented to balance the labeling cost and accuracy. Compared with binary mask and depth map, face position map is accurate and fast.
	\item We build a large and comprehensive dataset, in which live and attack samples are collected by different types of mobile ends in various scenarios.
	\item The effectiveness, performance, and advantages of the proposed method are numerically and experimentally verified.
\end{itemize}

\section{The Proposed Method}
\label{sec:method}

In this section, we first introduce a temporal feature fusion network based on eulerian magnification (EulerNet), including feature-compressed attention module (FCAM) and Residual Pyramid. Following EulerNet, a lightweight face labeling method based on face landmarks is described and the supervision method based on face location map is presented.

\subsection{EulerNet}
\label{subsec:net}
\begin{figure}[htb]
	\centering
	\subfloat[Live]{
		\includegraphics[width=0.3\linewidth]{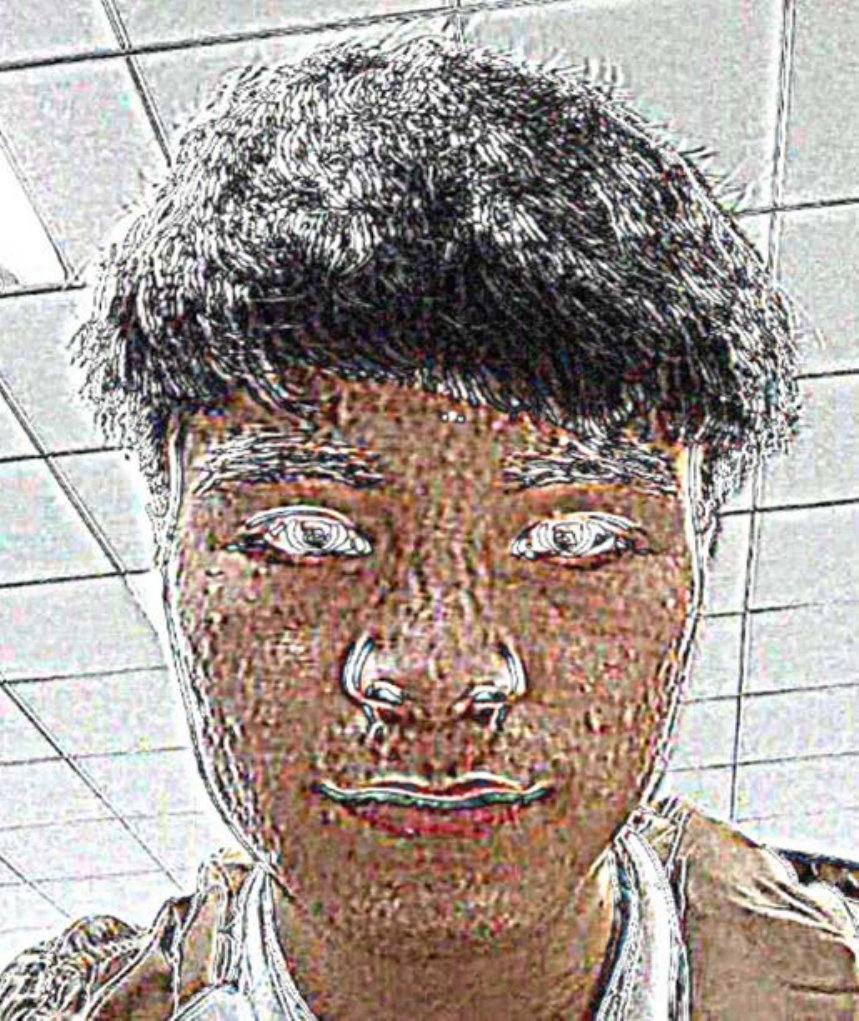}
		\label{}}
	\hfil
	\subfloat[Print]{\includegraphics[width=0.3\linewidth]{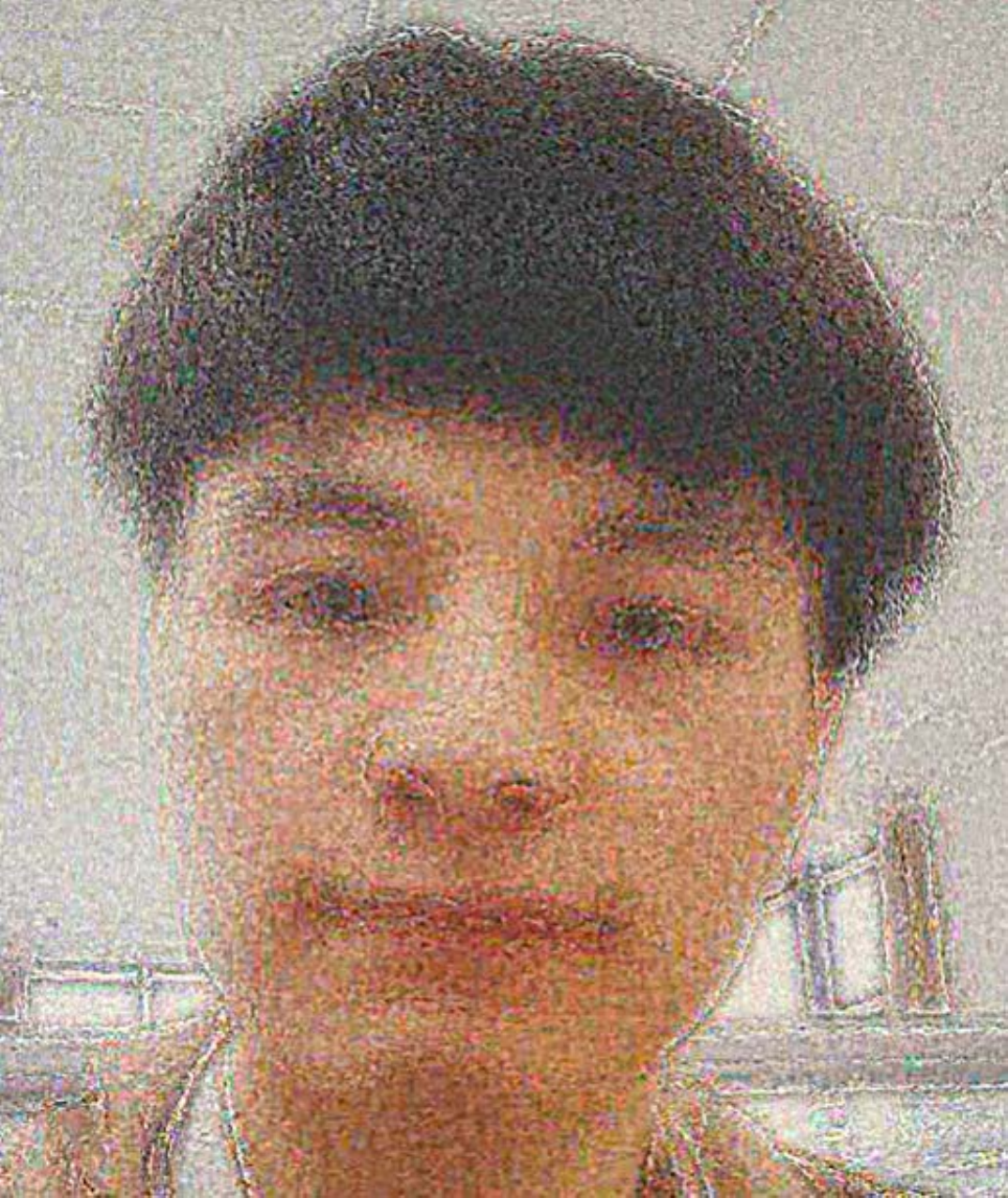}
		\label{}}
	\hfil
	\subfloat[Replay]{\includegraphics[width=0.3\linewidth]{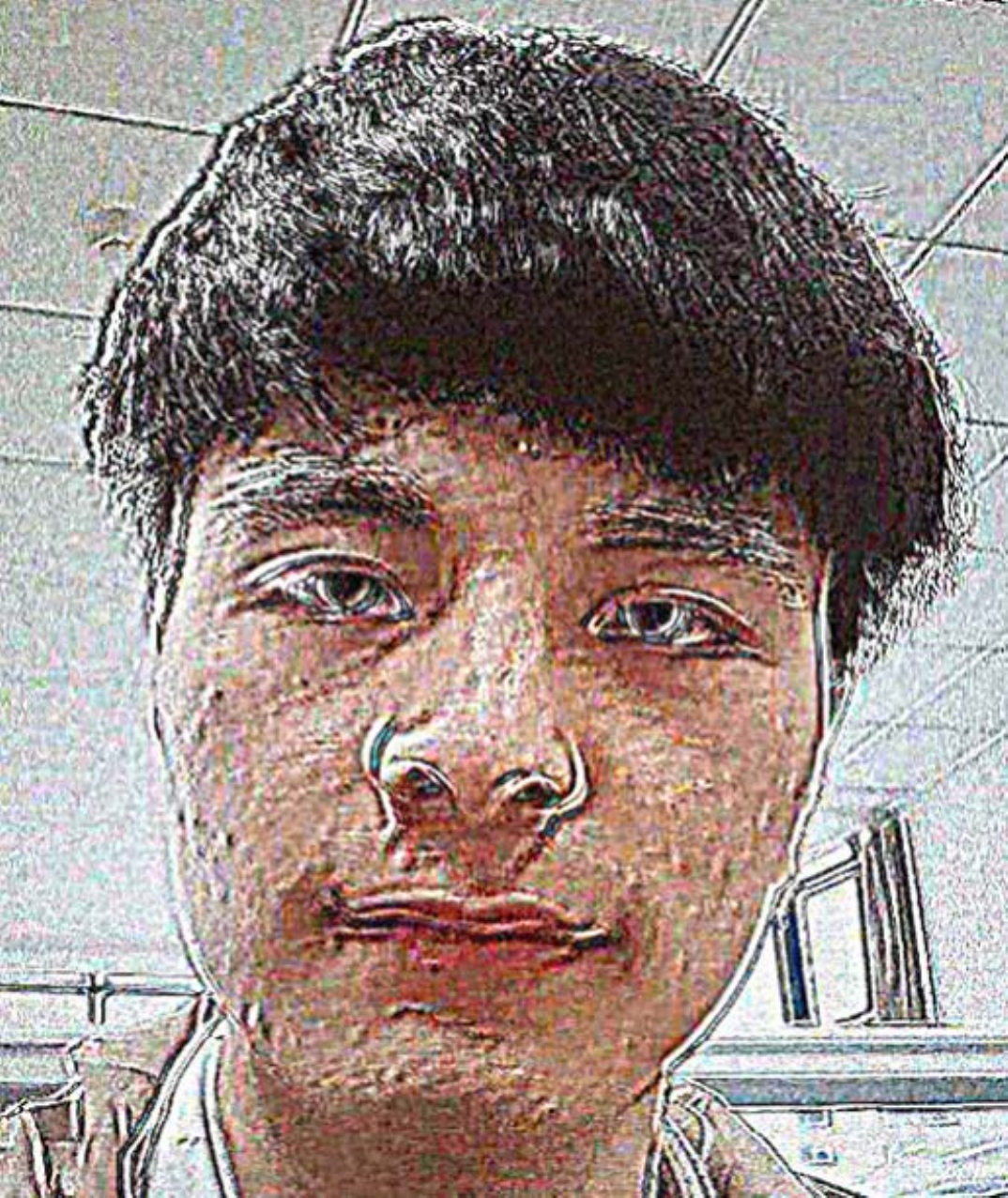}
		\label{}}
	\hfil
	\caption{Results of Applying Eulerian Video Magnification}
	\label{fig:eulerCompare}
\end{figure}

\begin{figure}[t]
	\centering
	\includegraphics[width=0.8\linewidth]{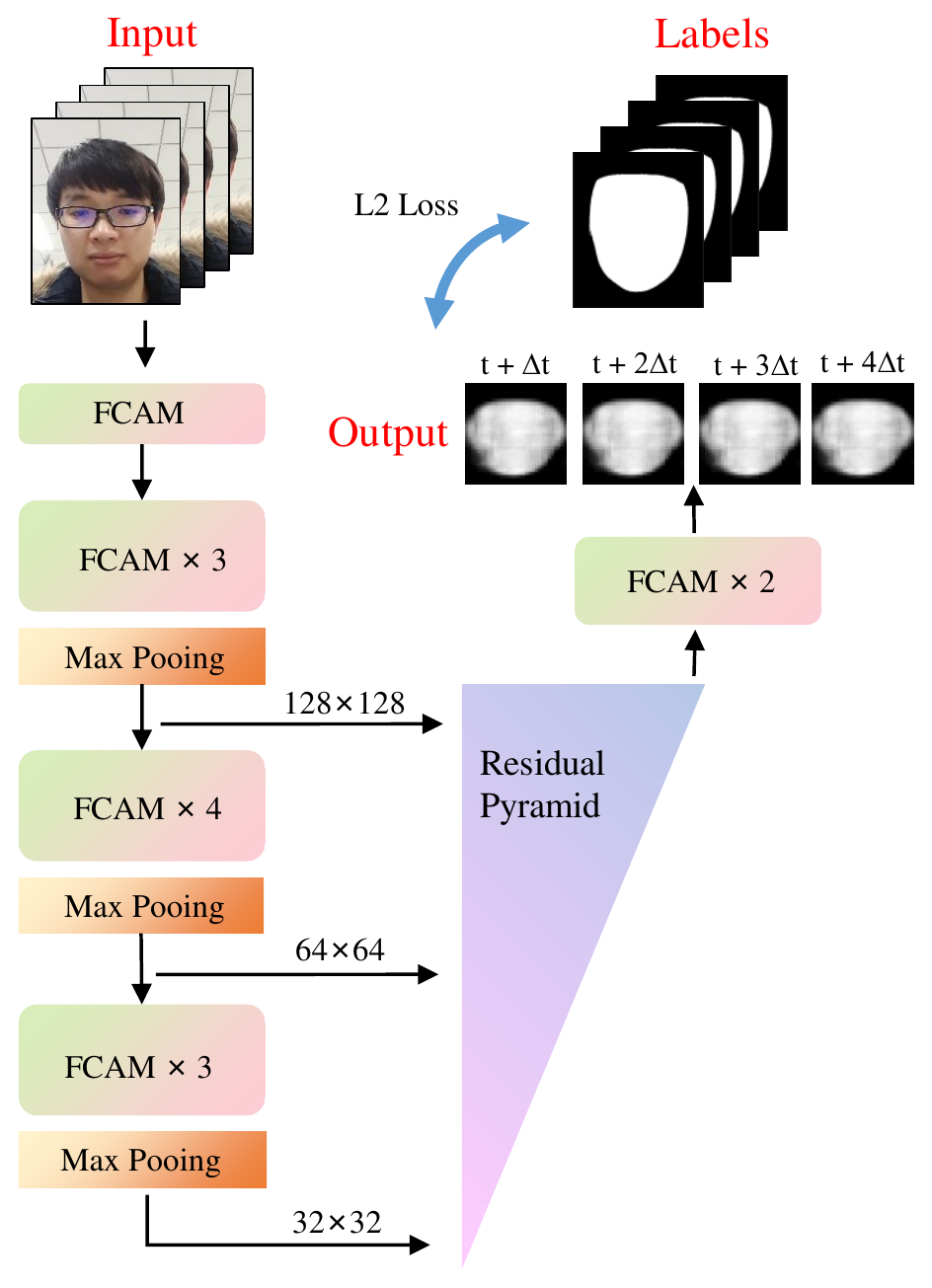}
	\caption{The proposed EulerNet.}
	\label{fig:network}
\end{figure}

We propose a temporal feature fusion network based on eulerian magnification, namely EulerNet, to extract the abnormal clues from live and spoofing. By applying eulerian video magnification \cite{wu2012eulerian} to live and spoofing faces, the import clues for face anti-spoofing are discovered. As illustrated in \reffig{fig:eulerCompare}, the result of print attack sample, lack a natural motion transition has obvious distinguishability with other attacks. Although replay attack and live sample have the same motion transition, the pattern characteristics are different, which provides reliable information for the network. \reffig{fig:network} shows EulerNet architecture includes a Residual Pyramid and a series of feature-compressed attention modules (FCAM) and max pooling. Instead of processing the whole video, we extract a sequence (length 4 and frame interval 3) from the video as input to avoid the problems of excessive memory usage and high derivation time overhead. The attention structure was designed in FCAM to process the temporal features. Using differential infinite impulse response filtering, FCAM amplify the subtle changes in faces between different frames. In \reffig{fig:network}, the feature level is divided from 256×256 into 128×128, 64×64, and 32×32 in EulerNet, which is fused by Residual Pyramid. Different levels of features (128×128, 64×64, and 32×32) are sampled for residuals to obtain effective frequency for face anti-spoofing.

\begin{figure}[htb]
	\centering
	\includegraphics[width=0.99\linewidth]{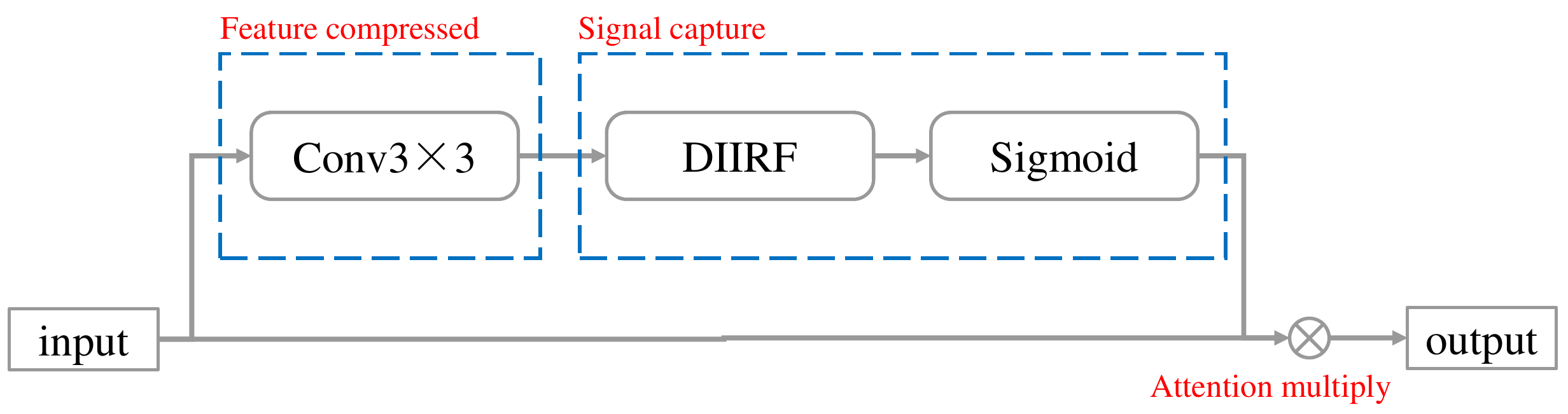}
	\caption{Feature-compressed attention module.}
	\label{fig:attention}
\end{figure}

\textbf{Feature-compressed Attention Module}. In FCAM, the feature map channels are adjusted to 1 by the 3×3 convolution (Conv3×3), which synthesizes information from each channel. Inspired by the realization of speech signal \cite{kuznetsov2020differentiable}, we transform the implementation of an infinite impulse response (IIR) filter on the image feature map to construct a differential infinite impulse response filter (DIIRF). By initializing a state matrix at the feature map size with 0, the output and the update for state in time sequence can be described as

\begin{equation}
	\label{dif_iir_filters_1}
	{y\lbrack n\rbrack=b_0x\lbrack n\rbrack+h_1\lbrack n-1\rbrack}
\end{equation}
\begin{equation}\\
	\label{dif_iir_filters_2}
	{h_1\lbrack n\rbrack=b_1x\lbrack n\rbrack+h_2\lbrack n-1\rbrack-a_1y\lbrack n\rbrack}
\end{equation}
\begin{equation}\\
	\label{dif_iir_filters_3}
	{h_2\lbrack n\rbrack=b_2x\lbrack n\rbrack-a_2y\lbrack n\rbrack}
\end{equation}
where ${y\lbrack n\rbrack}$ and ${x\lbrack n\rbrack}$ present output and input at nth timestamp, respectively. ${h_i}$ is the state matrix with 0. ${b_i}$ and ${a_j}$ is the training parameters of the filter layer, ${i\in\left[0,\;2\right]}$, ${j\in\left[1,\;2\right]}$. The sequential feature is utilized to filter video signals in DIIRF and the effective frequency for face anti-spoofing is reserved. The sigmoid function as the gate control unit process the feature from DIIRF. Attentional structure is built by multiplying the feature map obtained by sigmoid back to the original input. FCAM architecture is shown in \reffig{fig:attention}.

\textbf{Residual Pyramid}. To amplify the weak signal for face anti-spoofing, we design Residual Pyramid to fuse feature. As shown in \reffig{fig:network}, 3-level outputs from backbone network are collected as the input of Residual Pyramid. In Residual Pyramid, the 32×32 (${F_{32\times32}}$) and 64×64 (${F_{64\times64}}$) features are upsampled to generate 64×64 (${F_{64\times64}'}$) and 128×128 ${F_{128\times128}'}$) feature. Then, the ${F_{64\times64}'}$ and ${F_{128\times128}'}$ feature are subtracted by ${F_{64\times64}}$ and ${F_{128\times128}}$, respectively. After Conv3×3 extracting, the output ${S_{64\times64}}$ and ${S_{128\times128}}$ are obtained.
\begin{equation}
	\label{residual_2}
	{S_{128\times128}=Conv\left(F_{128\times128}-Upsample\left(F_{64\times64}\right)\right)}
\end{equation}
\begin{equation}
	\label{residual_1}
	{S_{64\times64}=Conv\left(F_{64\times64}-Upsample\left(F_{32\times32}\right)\right)}
\end{equation}
\begin{equation}
	\label{residual_3}
	\begin{split}
		Ouput_{32\times32}=Concat(
		& Donwsample(S_{128\times128}),  \\
		& Donwsample(S_{64\times64}), \\
		& F_{32\times32})
	\end{split}
\end{equation}
The Residual Pyramid can help the network directly learn useful information from features of different depths. In particular, upsampling and downsampling are introduced to calculate the residual between different resolutions. 

\begin{figure}[htb]
	\centering
	\includegraphics[width=1\linewidth]{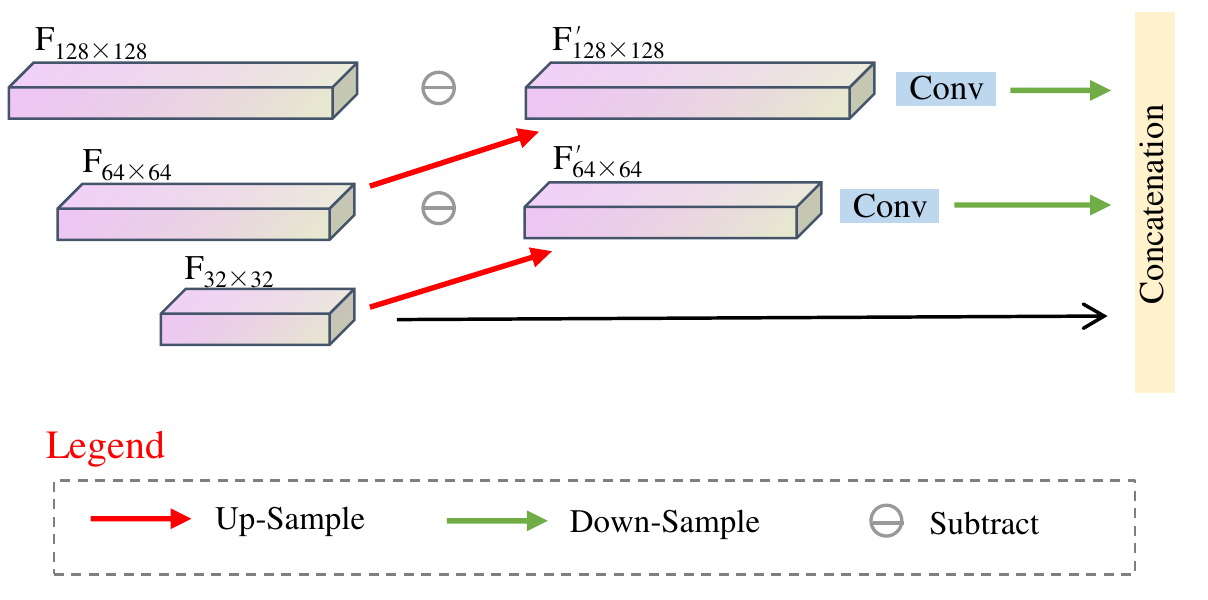}
	\caption{Residual Pyramid for feature fusing of different depths.}
	\label{fig:iir}
\end{figure}

\subsection{A lightweight face labeling method based on face landmarks}
\label{subsec:Segmentation Map}

\begin{figure}[htb]
	\centering
	\includegraphics[width=1\linewidth]{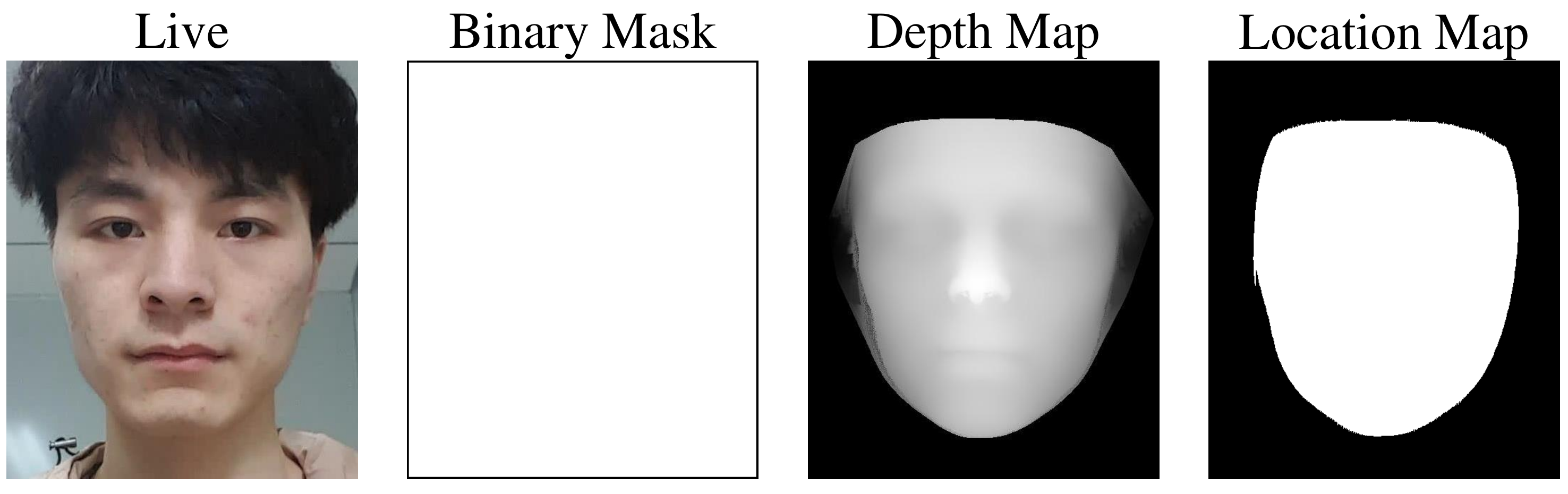}
	\caption{Comparison of different supervision for FAS.}
	\label{fig:map}
\end{figure}

Binary mask-based and Depth map-based labeling methods are effective, but both have drawbacks. Binary mask \cite{george2019deep} does not distinguish between the face and background, and discards the depth information, which reduces the network efficiency. For depth map-based labeling methods, additional labeling resources are required. Compared with binary mask-based and Depth map-based labeling methods, we connect the key points of the face contour to form a closed face region, namely face position map. In particular, the pixel points within the face region are labeled as 1 and 0 on live face and spoofing, respectively. Face position map is used to reduce the labeling time, preserve the gradient information of the face edges, and improve accuracy. \reffig{fig:map} shows the binary mask, depth map, and face position map.

\section{Dataset Collection}
\label{sec:dataset}

The difference of data distribution in different application scenarios determines the model performance in a specific scenario. To address this problem, we simulate the real-world scenario to establish the face anti-spoofing dataset, where each data sample is captured by the mobile end camera. In our dataset, the majority of mobile ends currently on the market are used for data collection. The spoofing type is divided into three parts: complex attack, certificate attack, and image dynamic generation attack. In particular, it is difficult to create and conduct 3D attacks in practical scenarios, so we mainly collect 2D attacks used in \cite{li2019face}. 

In our dataset, the complex attack includes various paper attacks, projection attacks, high-definition quality shots, and low-quality shots. Certificate attack is common in real-world scenarios and different from other attacks, which consists of ID card attacks, passport document attacks, etc. Image dynamic generation attack is an AI attack method, where dynamic images are generated from single images for face spoofing. 

\begin{figure}[htb]
	\centering
	\includegraphics[width=1\linewidth]{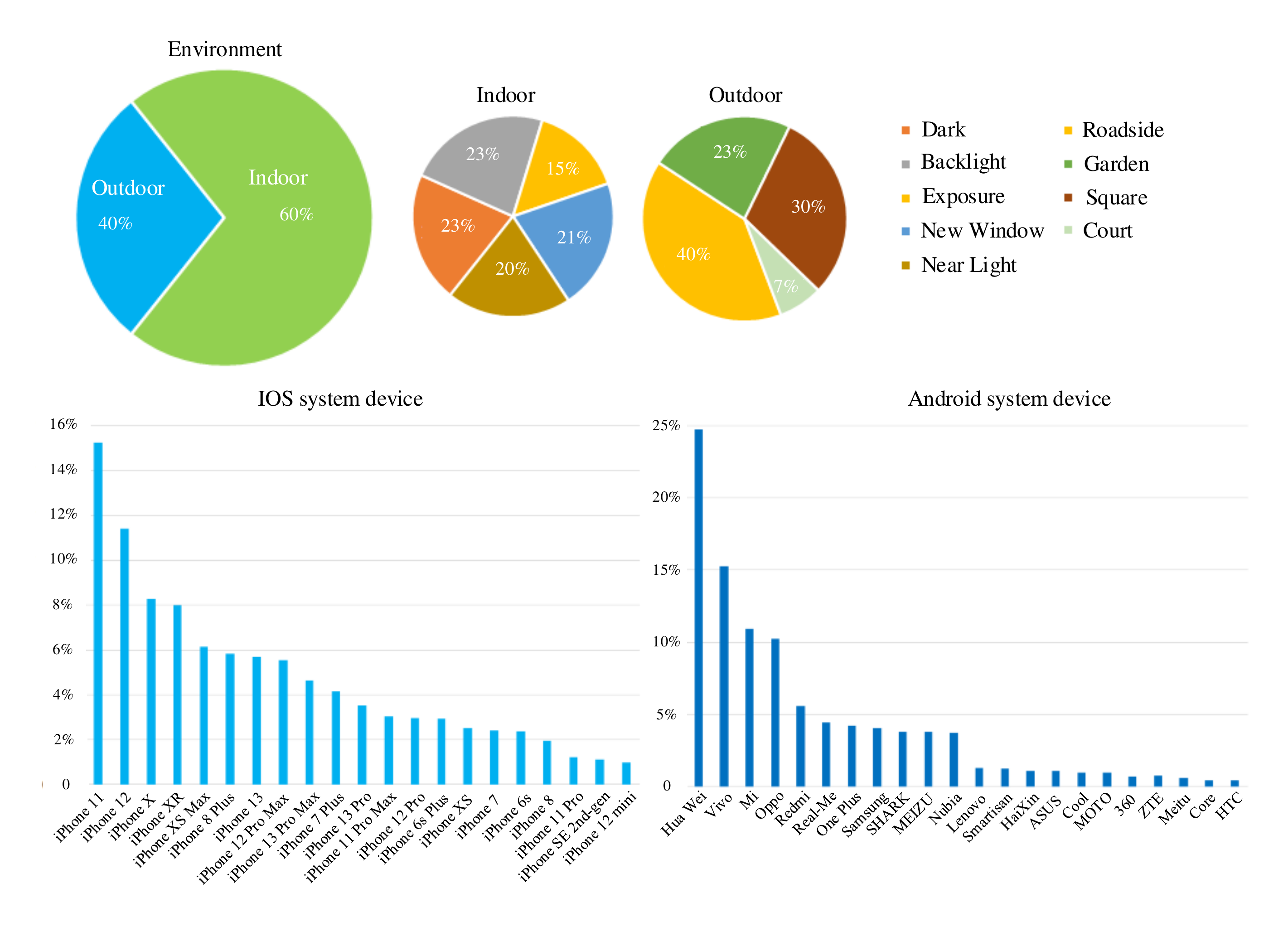}
	\caption{Statistics of electronic devices and scenes of our dataset.}
	\label{fig:dataimage}
\end{figure}

Our dataset outperforms previous public datasets \cite{chingovska2012effectiveness,chan2017face,zhang2012face} in three points: 1) Our dataset is the largest one that includes 30,000 live and spoofing videos (average duration to be 2s), collected from 10000 subjects, compared to 12,000 videos from 200 subjects in \cite{liu2019aurora}. 2) As shown in \reffig{fig:dataimage}, our dataset covers dozens of commonly used mobile ends. 3) The data distribution of our dataset matches the real-world mobile verification scenarios.

\section{Experiments}
\label{sec:exper}

\subsection{Implementation Details}
\label{subsec:detais}

To verify the effectiveness of our proposed network architecture and supervision method, we conduct experiments on the collected dataset, dividing the training, development, and test sets according to the ratio of 3:1:1. The learning rate is initialized as 1e-4 and the batchsize is set to 16 for a synergistic optimizer Ranger \cite{Ranger}. The input to the model are sequences of 4 consecutive frames taken from the same video at intervals of 3. Thus we compute 64 images in one gradient update. The model is trained with Nvidia Tesla V100s. Pytorch is utilized as the backend for the network architecture. L2 loss is utilized to guide the network predict the location map.

Average Classification Error Rate (ACER) is the mean value of Attack Presentation Classification Error Rate (APCER) and Bona Fide Presentation Classification Error Rate (BPCER), given by
\begin{equation}
	\label{acper_bpcer}
	{APCER=\frac{FP}{TN+FP},\;\;BPCER=\frac{FN}{TP+FN}}
\end{equation}
\begin{equation}
	\label{acer}
	{ACER=\frac{APCER+BPCER}2}
\end{equation}
In the next subsections, ACER is uniformly utilized for evaluation. For each video, we extracted multiple non-overlapping sequences, then calculated the live score individually and finally average scores for video.

\subsection{Ablation Study}
\label{subsec:ablation}

To investigate the behavior of EulerNet, we conducted several ablation studies. \reftab{ablation} shows the ablation study on our dataset. The baseline model is the proposed method using all improvements. Compare 1 (C1) and Compare 5 (C5) discuss the influence of different labels, including binary mask and depth map. In Compare 3 (C3), the Residual Pyramid is replaced by channels concatenation. Compare 4 (C4) removes the FCAM. Compare 2 (C2) discards all structure improvements and uses the common depth map for global comparison.

\textbf{Effectiveness of FCAM and Residual Pyramid}. After adding FCAM and d Residual Pyramid, ACER decreased by 0.34 percent and 0.38 percent, respectively, indicating that they were beneficial in FAS. The results suggest that extracting the effective frequency information of the input and fusing different resolution features together in a residual manner is meaningful for face anti-spoofing. 
\reftab{ablation} demonstrate that we must guide the network to mine as many anomalous patterns as possible in the attack video. The ACER of the proposed model is reduced by 0.7\% concerning Compare 2 in \reftab{ablation}. We optimize the combination of models and labels to maximize the benefits and obtain excellent accuracy on our dataset.

\textbf{Effectiveness of Face Location Map Supervision}. The optimal model on the development set is utilized to check the test set. Our proposed model yields the best ACER, achieving 0.18\% lower than the model supervised with depth map and 0.96\% lower than the model supervised with binary mask. This indicates that the features with binary mask supervision are not robust and discriminative, which are learned from labeling all pixels of living images as 1. The practice of regression on face location map reduces the difficulty of pixel modeling task and makes the network focus more on the color texture of face region and edge gradient information.

\begin{table}[tb]
	\centering
	\caption{Ablation study on our dataset.}
	\label{ablation}
	\begin{tabular}{|l|lcc|cc|}
		\hline
		\multicolumn{1}{|c|}{} &
		\multicolumn{3}{c|}{Structure} &
		\multicolumn{2}{c|}{{ACER(\%)$\downarrow$}} \\ \cline{2-6} 
		\multicolumn{1}{|c|}{\multirow{-2}{*}{Tag}} &
		\multicolumn{1}{c|}{Label} &
		\multicolumn{1}{c|}{{\begin{tabular}[c]{@{}c@{}}FCAM\end{tabular}}} &
		{\begin{tabular}[c]{@{}c@{}}Residual\\ Pyramid\end{tabular}} &
		\multicolumn{1}{c|}{{Dev}} &
		Test \\ \hline
		Compare 1 &
		\multicolumn{1}{l|}{Binary Mask} &
		\multicolumn{1}{c|}{{\color[HTML]{009901} $\surd$}} &
		{\color[HTML]{009901} $\surd$} &
		\multicolumn{1}{c|}{3.95} &
		2.84 \\ \hline
		Compare 2 &
		\multicolumn{1}{l|}{Depth Map} &
		\multicolumn{1}{c|}{{\color[HTML]{FE0000} $\times$}} &
		{\color[HTML]{FE0000} $\times$} &
		\multicolumn{1}{c|}{3.62} &
		2.57 \\ \hline
		Compare 3 &
		\multicolumn{1}{l|}{Face Location Map} &
		\multicolumn{1}{c|}{{\color[HTML]{009901} $\surd$}} &
		{\color[HTML]{FE0000} $\times$} &
		\multicolumn{1}{c|}{2.85} &
		2.26 \\ \hline
		Compare 4 &
		\multicolumn{1}{l|}{Face Location Map} &
		\multicolumn{1}{c|}{{\color[HTML]{FE0000} $\times$}} &
		{\color[HTML]{009901} $\surd$} &
		\multicolumn{1}{c|}{3.13} &
		2.22 \\ \hline
		Compare 5 &
		\multicolumn{1}{l|}{Depth Map} &
		\multicolumn{1}{c|}{{\color[HTML]{009901} $\surd$}} &
		{\color[HTML]{009901} $\surd$} &
		\multicolumn{1}{c|}{2.74} &
		2.06 \\ \hline
		Baseline &
		\multicolumn{1}{l|}{Face Location Map} &
		\multicolumn{1}{c|}{{\color[HTML]{009901} $\surd$}} &
		{\color[HTML]{009901} $\surd$} &
		\multicolumn{1}{c|}{\textbf{2.48}} &
		\textbf{1.88} \\ \hline
	\end{tabular}
\end{table}

\begin{figure}[t]
	\centering
	\includegraphics[width=0.95\linewidth]{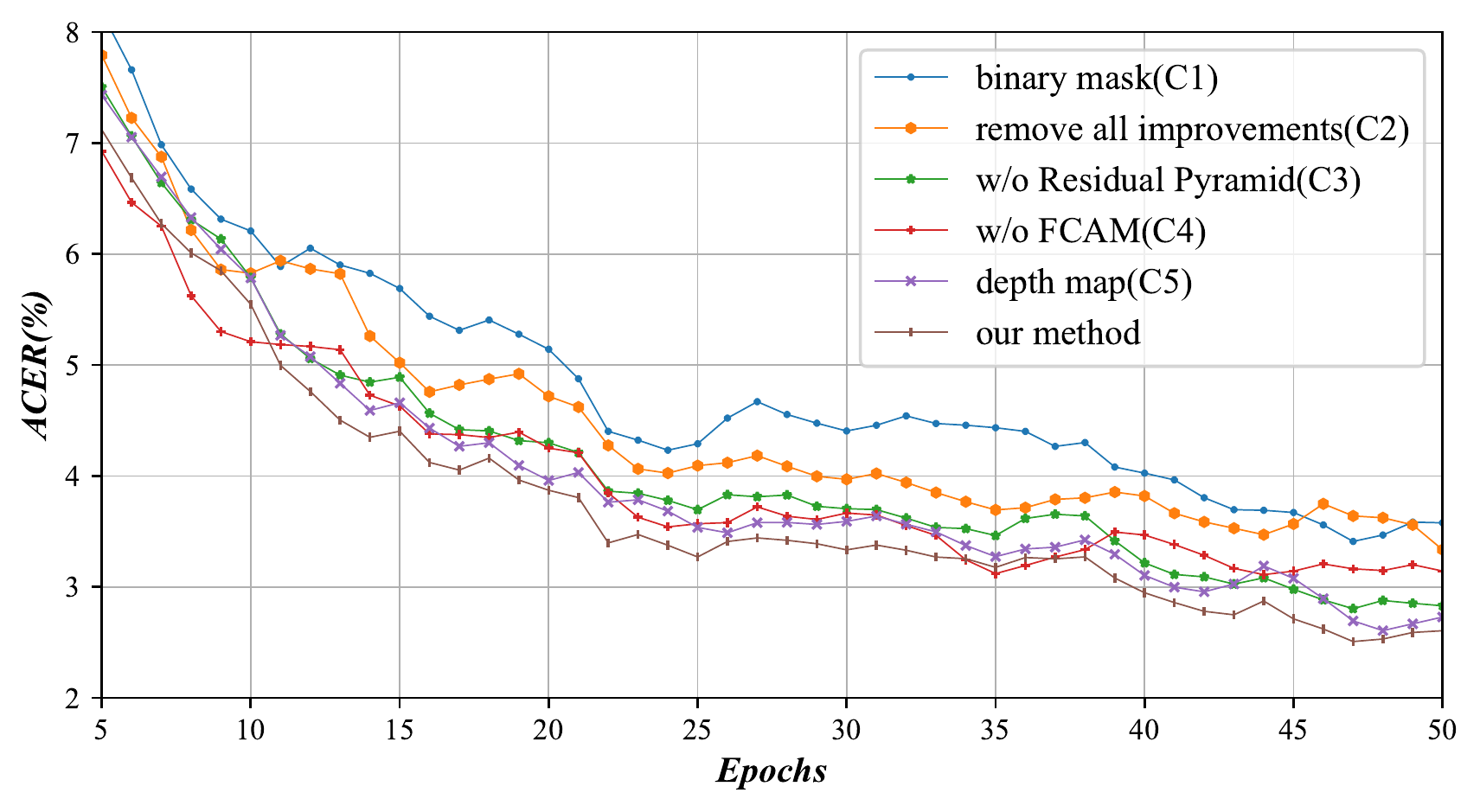}
	\caption{Performance comparison curve on the development set in training process.}
	\label{fig:curve}
\end{figure}

As shown in the \reffig{fig:curve}, we plot the training performance curves of different models. At the beginning, the curve without FCAM drops the fastest because the model does not extract frequency information. Although the early iteration is fast, the accuracy in the later stage is low, which suggests that it is significant to mine the specific frequency information. In the late training phase (epochs 45-50), the fluctuating curves are sorted by mean value size as C2$>$C1$>$C4$>$C3$>$C5$>$our method. Especially, the proposed method curve shows a smoother decreasing trend during training with less fluctuation.

\begin{figure}[tbp]
	\centering
	\includegraphics[width=0.96\linewidth]{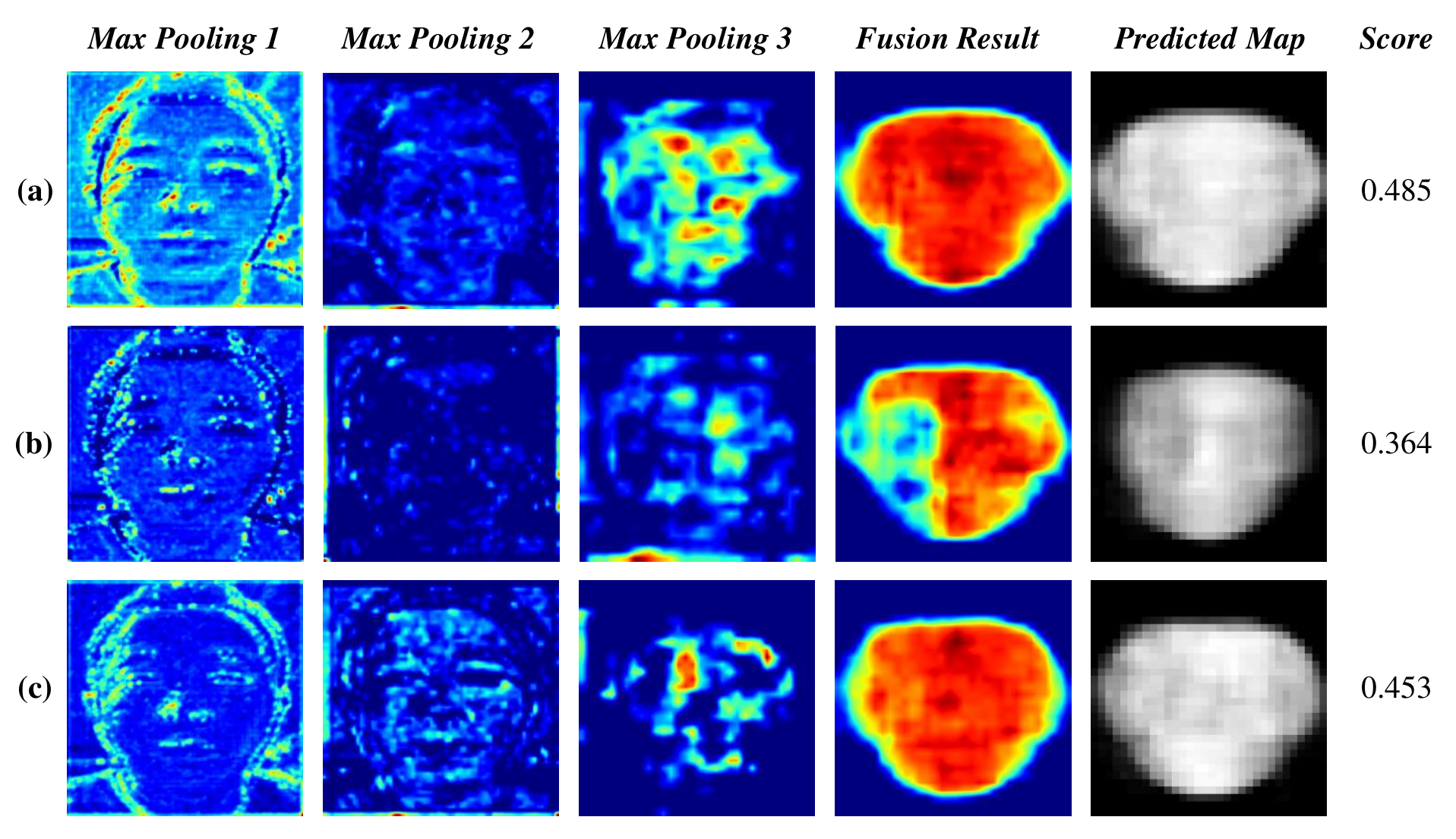}
	\caption{Visualization study of neuron outputs at different depths: (a) our method, (b) w/o FCAM, (c) w/o Residual Pyramid. From left to right are the results with increasing downsampling multiplier as well as the multi-resolution fusion results and the final prediction maps. Their resolutions are 128×128, 64×64, 32×32, 32×32, 32×32 in order.}
	\label{fig:compare2}
\end{figure}
\begin{figure}[t]
	\centering
	\includegraphics[width=0.9\linewidth]{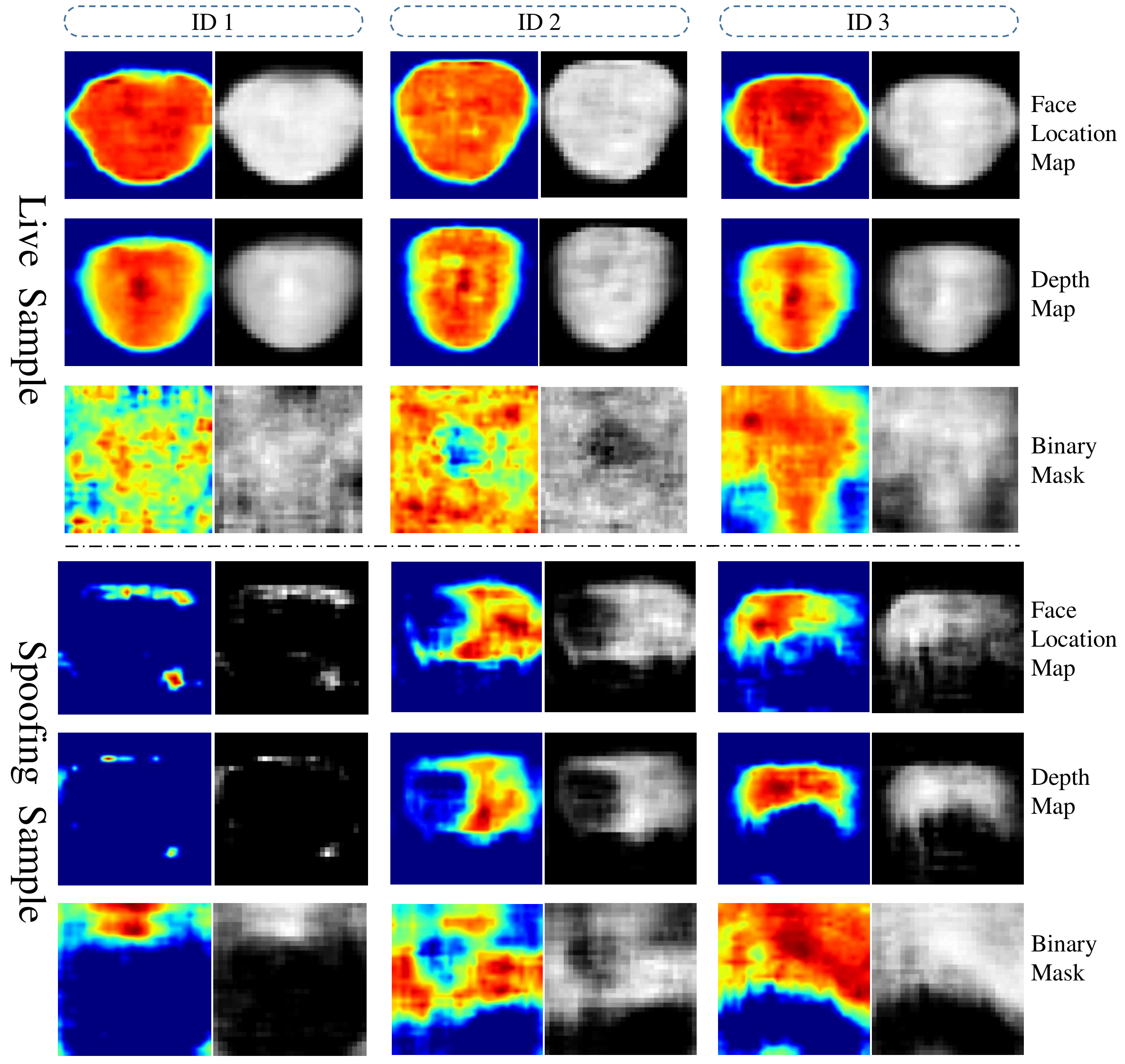}
	\caption{Model output and heatmap visualization under different supervision methods. The same column  indicates the results of the same face, marked with ID 1-3. }
	\label{fig:compare1}
\end{figure}
\begin{table}[t]
	\centering
	\caption{The results of intra testing on four protocols of OULU-NPU compared with deep learning methods}
	\label{oulu}
	\begin{tabular}{|c|c|c|c|c|} 
		\hline
		\multicolumn{1}{|l|}{Prot.~} & Method                    & APCER(\%)          & BPCER(\%)          & ACER(\%)           \\ 
		\hline
		\multirow{5}{*}{1}           & Disentangled \cite{zhang2020face}              & 1.7                & 0.8                & 1.3                \\ 
		\cline{2-5}
		& FAS-SGTD \cite{wang2020deep}                  & 2.0                & \textbf{0.0}     & 1.0                \\ 
		\cline{2-5}
		& DeepPixBiS \cite{george2019deep}                & 0.8                & \textbf{0.0}      & \textbf{0.4}      \\ 
		\cline{2-5}
		& CDCN \cite{yu2020searching}                      & \textbf{ 0.4 }     & 1.7                & 1.0                \\ 
		\cline{2-5}
		& \textbf{ EulerNet(Ours) } & \textbf{ 0.4 }     & 3.3                & 1.9                \\ 
		\hline
		\multirow{5}{*}{\textbf{2}}  & DeepPixBiS \cite{george2019deep}                & 11.4               & \textbf{ 0.6 }     & 6.0                \\ 
		\cline{2-5}
		& Disentangled \cite{zhang2020face}             & \textbf{1.1}     & 3.6                & 2.4                \\ 
		\cline{2-5}
		& FAS-SGTD \cite{wang2020deep}                  & 2.5                & 1.3                & 1.9                \\ 
		\cline{2-5}
		& CDCN \cite{yu2020searching}                      & 1.5                & 1.4                & \textbf{ 1.5 }     \\ 
		\cline{2-5}
		& \textbf{ EulerNet(Ours) } & 2.1                & 1.4                & 1.7                \\ 
		\hline
		\multirow{5}{*}{\textbf{3}}  & DeepPixBiS \cite{george2019deep}                & 11.7±19.6~         & 10.6±14.1          & 11.1±9.4           \\ 
		\cline{2-5}
		& FAS-SGTD \cite{wang2020deep}                  & 3.2±2.0            & 2.2±1.4            & 2.7±0.6            \\ 
		\cline{2-5}
		& CDCN \cite{yu2020searching}                      & \textbf{2.4±1.3} & 2.2±2.0            & 2.3±1.4            \\ 
		\cline{2-5}
		& Disentangled \cite{zhang2020face}              & 2.8±2.2            & 1.7±2.6            & 2.2±2.2            \\ 
		\cline{2-5}
		& \textbf{ EulerNet(Ours) } & 2.6±1.3            & \textbf{ 1.6±0.8 } & \textbf{2.1±0.5}  \\ 
		\hline
		\multirow{5}{*}{4}           & DeepPixBiS \cite{george2019deep}                & 36.7±29.7          & 13.3±14.1          & 25.0±12.7          \\ 
		\cline{2-5}
		& CDCN \cite{yu2020searching}                      & 4.6±4.6            & 9.2±8.0            & 6.9±2.9            \\ 
		\cline{2-5}
		& FAS-SGTD \cite{wang2020deep}                  & 6.7±7.5            & \textbf{3.3±4.1} & 5.0±2.2            \\ 
		\cline{2-5}
		& Disentangled \cite{zhang2020face}              & 5.4±2.9            & 3.3±6.0            & 4.4±3.0            \\ 
		\cline{2-5}
		& \textbf{EulerNet(Ours)} & \textbf{ 1.8±1.9 } & 4.3±2.4            & \textbf{3.1±0.9}  \\
		\hline
	\end{tabular}
\end{table}
\subsection{Visualization}
\label{subsec:visual}
The Grad-cam \cite{selvaraju2017grad}  is used to visualize the output of the neurons in the model to compare the changes brought by changing the structure or the supervision method. \reffig{fig:compare2} details the effect of FCAM and Residual Pyramid on the features extracted by the network. Max Pooling 1-3 indicates the results using different downsampling multipliers. Fusion result is the multi-resolution fusion results. The final prediction are shown in Predicted Map. Scores are presented in the last column of \reffig{fig:compare2}. The higher the score on the far right, the higher the likelihood of being judged as live faces. The comparison between Max Pooling 1 and Max Pooling 2 shows that the visualization features are clearly distinct under different downsampling scales. The model with FCAM pays more attention to the parts where the action occurs, so there are higher activation values at pixels of the contour, eyes, and nose. Finally, the face score obtained by our method is 0.485, which is better than the model removed FCAM and Residual Pyramid. In addition, as the network deepened, the extracted features mined more abnormal clues and produced more accurate face location maps.

\reffig{fig:compare1} shows the visualization results for the three different labels (face location map, binary map, and depth map) are studied in this paper. For the live face, the heat map supervised by the face localization map is sharper and finer at the edges of the face region. As shown in \reffig{fig:compare1}, the binary mask-based supervised method ignores the gradient information of face edges. Compared with the depth map, the prediction map based on the face location map has higher contrast in distinguishing faces and backgrounds. When the input is spoofing face, the predictions obtained by the face location map-based and depth map-based methods are similar and still bias the face region, which demonstrates that the network does not need to simulate depth information in practice. As shown in \reffig{fig:compare1}, the results of the binary mask have less distinct semantic features. The data distribution on the heat map is more uniform and scattered with the binary mask supervision. Compared with the above three supervision methods, the loss of generalization of the features extracted by the network without the drive of depth and location information.

\subsection{Comparison on Public Dataset}
\label{subsec:public}

We compare the proposed EulerNet with recent deep learning methods on OULU-NPU \cite{boulkenafet2017oulu}. Protocol 1-3 respectively evaluate the generalization of the algorithm under previously unseen environmental conditions, attacks created with different printers or displays, and input camera variation. Protocol 4 considers all the above three factors. Generally, protocols 3 and 4 are more difficult than other protocols.

The Comparison results on OULU-NPU are shown in \reftab{oulu}. The best performance obtained by the proposed method in protocols 3 and 4 demonstrates that our method can maintain accuracy under complex conditions. \reftab{oulu} proves that the structures we have designed are capable of extracting abnormal clues, which has a stronger generalization in facing changes of the shooting device. The complexity of protocols 3 and 4 is similar to the realistic scenario where electronic products are changing rapidly, there are demonstrates the practicality of our proposed method.

\section{Conclusion}
In this paper, we propose a novel face anti-spoofing method, which effectively recognize the subtle differences between real face and spoofing in the video. The novel network architecture, namely EulerNet, is designed to fuse temporal information and extract abnormal clues. We propose a lightweight labeling method based on face landmarks to reduce the labeling cost and improve the labeling speed. Extensive experimental results on our datasets and public OULU-NPU validate the effectiveness of our method.

\bibliographystyle{IEEEtran}
\bibliography{bibtex/seke}

\end{document}